\def\year{2021}\relax
\documentclass[letterpaper]{article} 
\usepackage{aaai21}  
\usepackage{times}  
\usepackage{helvet} 
\usepackage{courier}  
\usepackage[hyphens]{url}  
\usepackage{graphicx} 
\usepackage{natbib}  
\usepackage{caption} 

\usepackage{subcaption}
\usepackage{amsthm}
\usepackage{algorithm}
\usepackage{algorithmicx}
\usepackage{algpseudocode}
\usepackage{amsmath}
\usepackage{amssymb}
\usepackage{booktabs}
\usepackage{multirow}
\usepackage[section]{placeins}

\nocopyright
\title{Automatic Curriculum Learning With \\ Over-repetition Penalty for Dialogue Policy Learning}
\author{
    Yangyang Zhao,
    Zhenyu Wang \thanks{Corresponding authors.\\This paper has accepted by AAAI2021},
    Zhenhua Huang
    \\
}
\affiliations{
    School of software engineering, South China University of Technology\\

    msyyz@mail.scut.edu.cn, wangzy@scut.edu.cn, sezhhuangscut@mail.scut.edu.cn 

}

\begin{document}

\maketitle

\begin{abstract}
Dialogue policy learning based on reinforcement learning is difficult to be applied to real users to train dialogue agents from scratch because of the high cost. User simulators, which choose random user goals for the dialogue agent to train on, have been considered as an affordable substitute for real users. However, this random sampling method ignores the law of human learning, making the learned dialogue policy inefficient and unstable. We propose a novel framework, Automatic Curriculum Learning-based Deep Q-Network (ACL-DQN), which replaces the traditional random sampling method with a teacher policy model to realize the dialogue policy for automatic curriculum learning. The teacher model arranges a meaningful ordered curriculum and automatically adjusts it by monitoring the learning progress of the dialogue agent and the over-repetition penalty without any requirement of prior knowledge. The learning progress of the dialogue agent reflects the relationship between the dialogue agent's ability and the sampled goals' difficulty for sample efficiency. The over-repetition penalty guarantees the sampled diversity. Experiments show that the ACL-DQN significantly improves the effectiveness and stability of dialogue tasks with a statistically significant margin. Furthermore, the framework can be further improved by equipping with different curriculum schedules, which demonstrates that the framework has strong generalizability.
\end{abstract}

\section{Introduction}

\begin{figure*}[tbp]
\centering
\subcaptionbox{Policy learning with user simulators. \label{fig:a}}{\includegraphics[width=0.9\columnwidth]{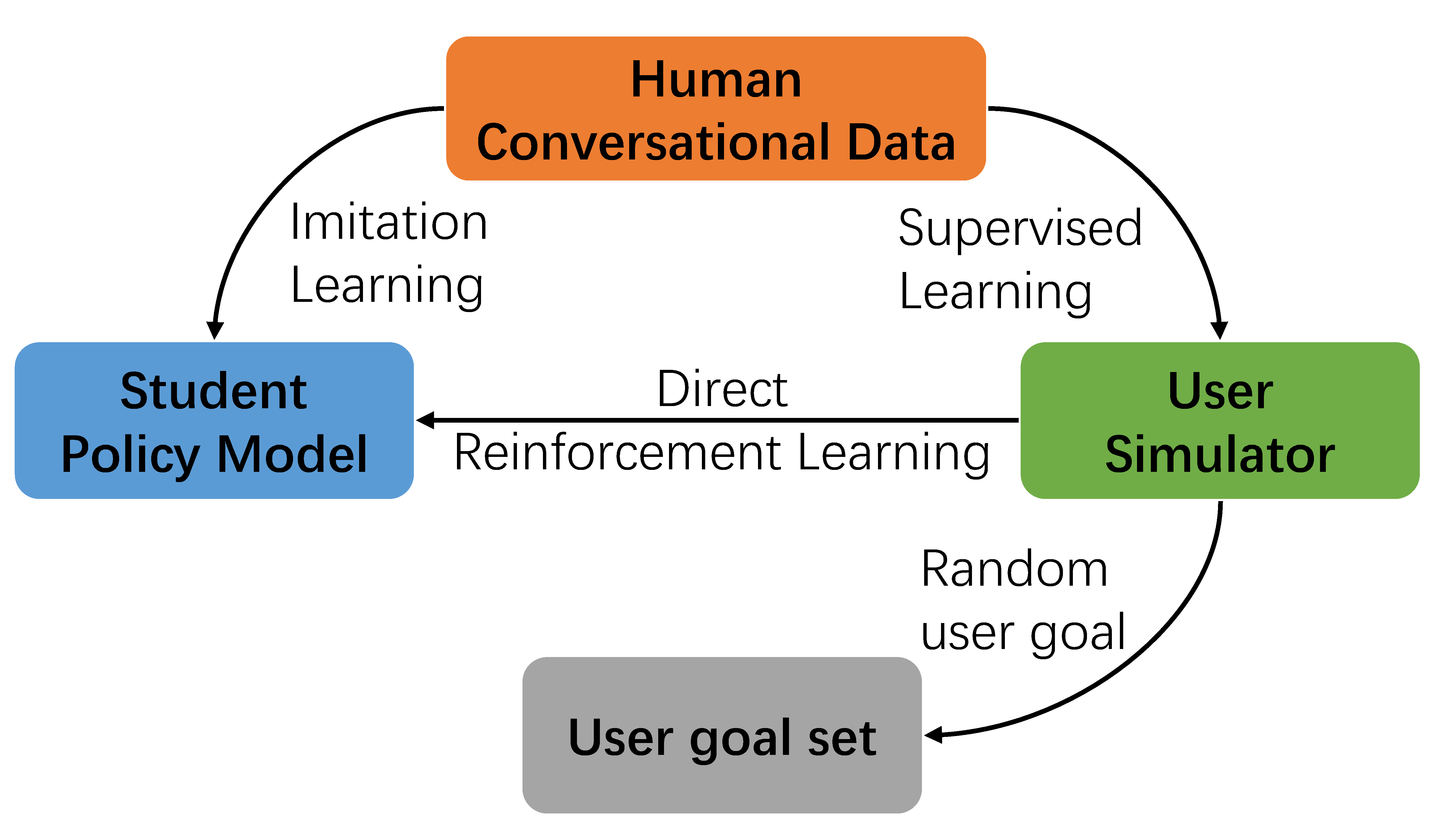}}
\subcaptionbox{Policy learning with proposed ACL-DQN framework. \label{fig:b}}{\includegraphics[width=0.9\columnwidth]{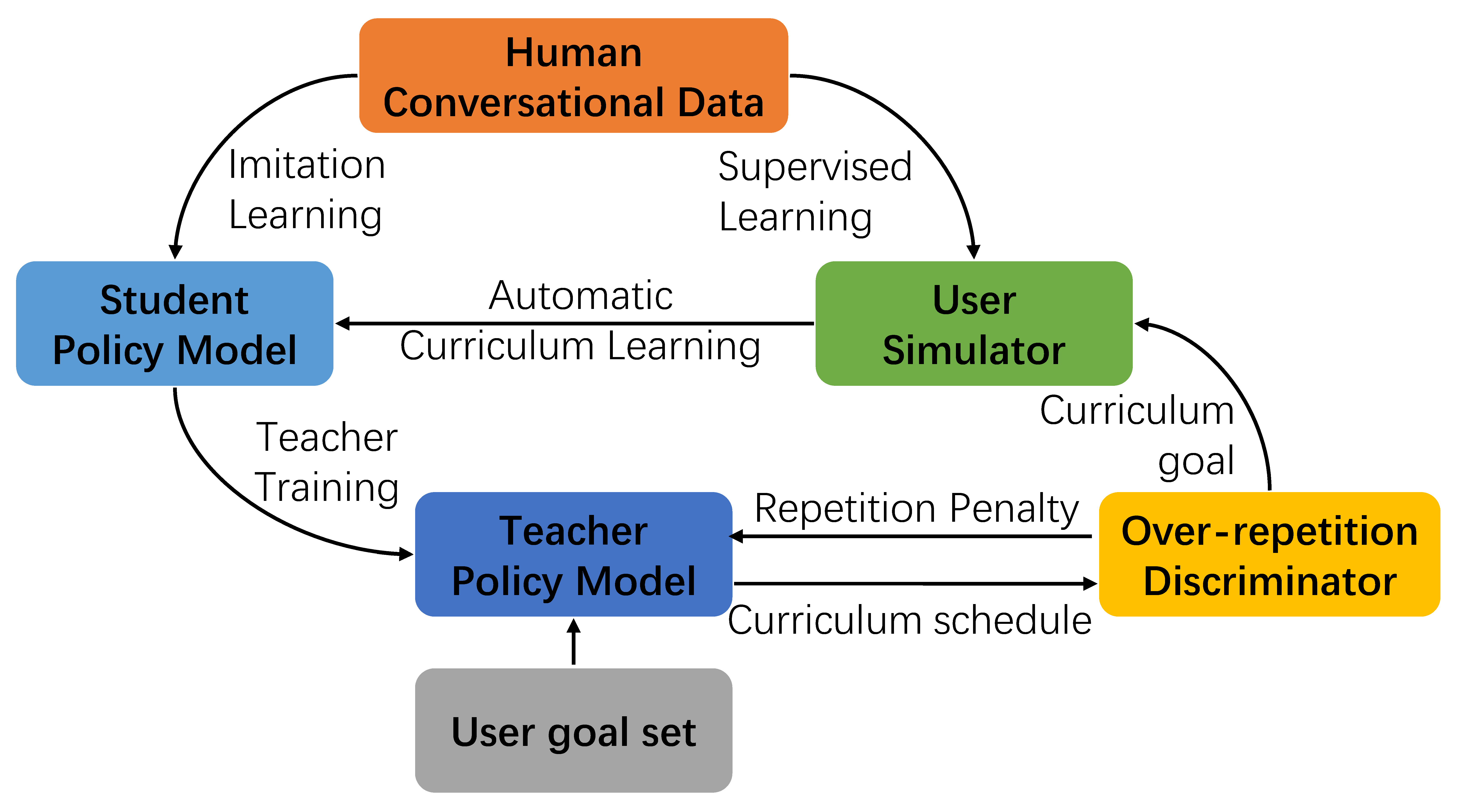}}
\caption{Two strategies of user simulator sampling for learning task-oriented dialogue policies via RL.}     
\label{fig:1} 
\end{figure*}

\noindent Learning dialogue policies are typically formulated as a reinforcement learning (RL) problem \cite{SuttonB98, YoungGTW13}. However, dialogue policy learning via RL from scratch in real-world dialogue scenarios is expensive and time-consuming, because it requires real users to interact with and adjusts its policies online
\cite{MnihKSRVBGRFOPB15,SilverHMGSDSAPL16,DhingraLLGCAD17,SuGMRUVWY16,LiCLGC17}. A plausible strategy is to use user simulators as an inexpensive alternative for real users, which randomly sample a user goal from the user goal set for the dialogue agent training \cite{SchatzmannTWYY07,SuGMRUVWY16a,LiCLGC17,BudzianowskiUSM17,PengLLGCLW17,LiuL17,PengLGLCW18}. In task-oriented dialogue settings, the entire conversation revolves around the sampled user goal implicitly. Nevertheless, the dialogue agent's objective is to help the user to accomplish this goal even though the agent knows nothing about this sampled user goal \cite{SchatzmannY09,LiLDLGC16}, as shown in Figure~\ref{fig:a}.

The randomly sampling-based user simulator neglects the fact that human learning supervision is often accompanied by a curriculum \cite{RenDLC18}. For instance, when a human-teacher teaches students, the order of presented examples is not random but meaningful, from which students can benefit \cite{BengioLCW09}. Therefore, this randomly sampling-based user simulators bring two issues:
\begin{itemize}
\item \textit{efficiency} issue: 
since the ability of the dialogue agent does not match the difficulty of the sampled user goal, it takes a long time for the dialogue agent to learn the optimal strategy (or fail to learn).
For example, in the early learning phase, it is possible that the random sampling method arranges the dialogue agent to learn more complex user goals first, and then learn simpler user goals. 
\item \textit{stability} issue: 
using random user goals to collect experience online is not stable enough, making the learned dialogue policy unstable and difficult to reproduce. Since RL is highly sensitive to the dynamics of the training process, dialogue agents trained with stable experience can guide themselves more effectively and stably than dialogue agents trained with instability.
\end{itemize}

Most previous studies of dialogue policy have focused on the \textit{efficiency} issue, such as reward shaping \cite{KulkarniNST16,LuZC19,ZhaoWYZHW20,CaoLCZ20}, companion learning \cite{ChenYCYZY17,ChenZCYY17}, incorporate planning \cite{GaoWPLL18,SuLGLC18,WuLLGY19,ZhaoWYZHW20,ZhangCSWD20}, etc.
However, \textit{stability} is a pre-requisite for the method to work well in real-world scenarios.
It is because, no matter how effective an algorithm is, an unstable online leaned policy may be ineffective when applied in the real dialogue environment. This can lead to bad user experience and thus fail to attract sufficient real users to continuously improve the policy. As far as we know, little work has been reported about the stability of dialogue policy. Therefore, it is essential to address the stability issue.

In this paper,  we propose a novel policy learning framework that combines curriculum learning and deep reinforcement learning,  namely Automatic Curriculum Learning-based Deep Q-Network (ACL-DQN).
As shown in Figure~\ref{fig:b}, this framework replaces the traditional random sampling method in the user simulator with a teacher policy model that arranges a meaningful ordered curriculum and dynamically adjusts it to help dialogue agent (also referred to student agent in this paper) for automatic curriculum learning. As a scheduling controller for student agents, the teacher policy model arranges students to learn different user goals in different learning stages without any requirement of prior knowledge. Sampling the user goals that match the ability of student agents regarding different difficulty of each user goal, can not only increases the feedback of the environment to the student agent but also makes the learning of the student agent more stable.

There are two criteria for evaluating the sampling order of each user goal: the learning progress of the student agent and the over-repetition penalty. The learning progress of the student agent emphasizes the efficiency of each user goal, encouraging the teacher policy model to choose the user goals that match the ability of the student agent to maximize the learning efficiency of the student agent. The over-repetition penalty emphasizes the sampled diversity, preventing the teacher policy model from \textit{cheating}\footnote[1]{The teacher policy model repeatedly selects user goals that the student agent has mastered to obtain positive rewards.}. The incorporation of the learning progress of the student agent and the over-repetition penalty reflects both sampled efficiency and sampled diversity to improve efficiency as well as stability of ACL-DQN. 

Additionally, the proposed ACL-DQN framework can equip with different curriculum schedules. Hence, in order to verify the generalization of the proposed framework, we propose three curriculum schedule standards for the framework for experimentation:
i) \textit{Curriculum schedule A}: there is no standard, only a single teacher model;
ii) \textit{Curriculum schedule B}: user goals are sampled from easiness to hardness in proportion;
iii) \textit{Curriculum schedule C}: ensure that the student agents have mastered simpler goals before learning more complex goals.

Experiments have demonstrated that the ACL-DQN significantly improves the dialogue policy through automatic curriculum learning and achieves better and more stable performance than DQN. Moreover, the ACL-DQN equipped with the curriculum schedules can be further improved. Among the three curriculum schedules we provided, the ACL-DQN under curriculum schedule C with the strength of supervision and controllability, can better follow up on the learning progress of students and performs best. In summary, our contributions are as follows:

\begin{itemize}
\item We propose Automatic Curriculum Learning-based Deep Q-Network (ACL-DQN). As far as we know, this is the first work that applies curriculum learning ideas to help the dialogue policy for automatic curriculum learning.
\item We introduce a new user goal sampling method (i.e., teacher policy model) to arrange a meaningful ordered curriculum and automatically adjusts it by minoring the learning progress of the student agent and the over-repetition penalty.
\item We validate the superior performance of  ACL-DQN by building dialogue agents for the movie-ticket booking task. The efficiency and stability of ACL-DQN are verified by simulation and human evaluations. Moreover, ACL-DQN can be further improved by equipping curriculum schedules, which demonstrates that the framework has strong generalizability.
\end{itemize}

\section{Proposed framework}

\begin{figure}
\centering
\subcaptionbox{Curriculum schedule A. \label{curriculum A}}
	{\includegraphics[width=0.9\columnwidth]{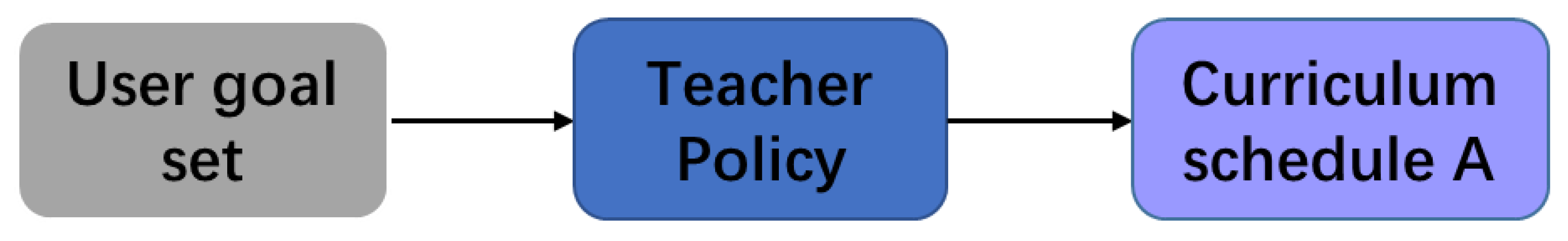}}\\[2ex]
\subcaptionbox{Curriculum schedule B. \label{curriculum B}}
	{\includegraphics[width=0.9\columnwidth]{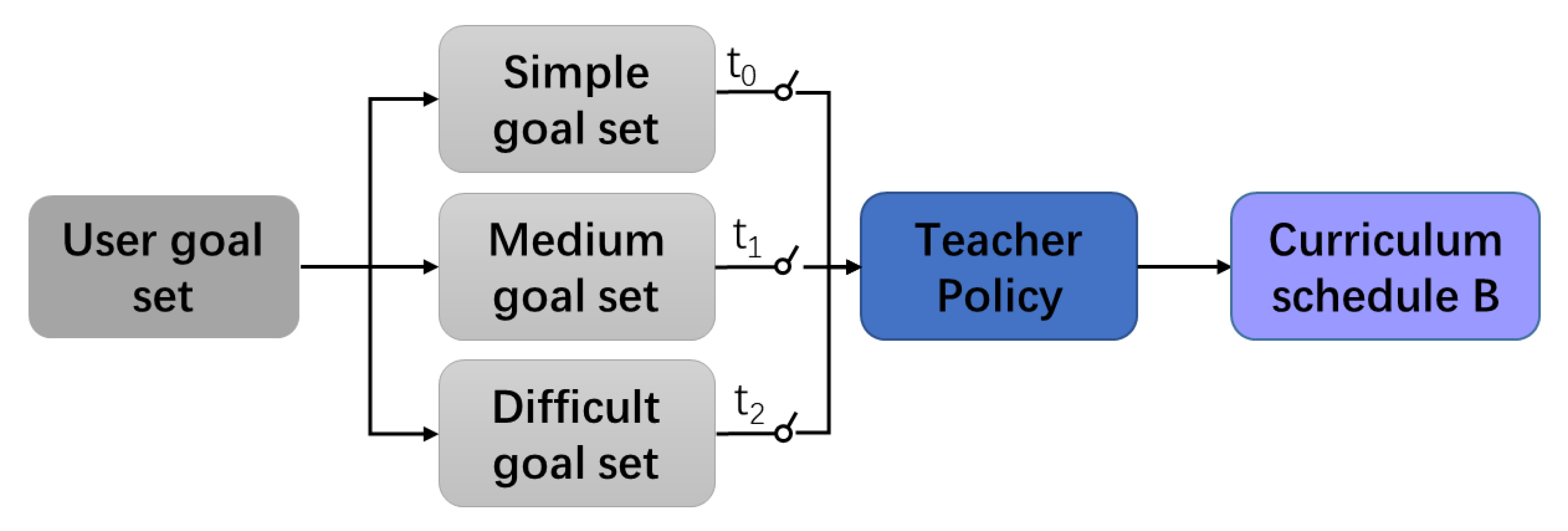}}\\[2ex]
\subcaptionbox{Curriculum schedule C. \label{curriculum C}}
	{\includegraphics[width=0.9\columnwidth]{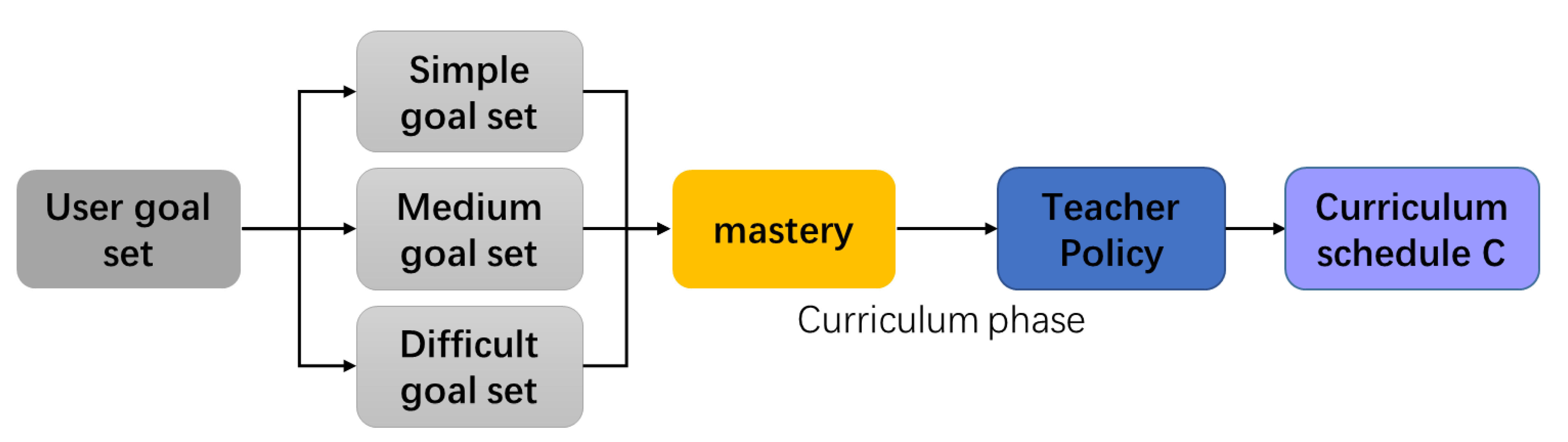}}
\caption{Three curriculum schedules were arranged and adjusted by the DQN-based teacher model based on three standards by monitoring student's training process and the over-repetition penalty (the feedback is shown in Figure~\ref{fig:a}).}     
\label{fig:curriculum}
\end{figure}

The proposed framework is illustrated in Figure~\ref{fig:a}, the ACL-DQN agent training consists of four processes:
(1) \textit{curriculum schedule} includes three strategies (Figure~\ref{fig:curriculum}), which are arranged by the teacher policy model based on three standards we provided and automatically adjusted according to the learning process of the student agent and the over-repetition penalty.
(2) \textit{over-repetition penalty}, which punishes the \textit{cheating} behaviors of the teacher policy model to guarantee the sampled diversity.
(3) \textit{automatic curriculum learning}, where the student agent interacts with a user simulator revolving around curriculum goal specified by the teacher policy model, collects experience, improves the student dialogue policy, and feeds its performances back to the teacher policy model for adjusting.
(4) \textit{teacher reinforcement learning}, where the teacher policy model is leaned and refined through a separate teacher experience replay buffer. 

\subsection{Curriculum schedule}

In this section, we introduce a DQN-based teacher model and three curriculum schedules, which are later used in the (2), (3), and (4) processes mentioned above.

\subsubsection{DQN-based teacher model}

The goal of the teacher model is to help the student agent learn a series of user goals sequentially. We can formalize the teacher goal as a Markov decision process (MDP) problem, which is well-suitable for reinforcement learning to solve:
\begin{itemize}
\item The state $s_t$ consists of five components: 1) the state provided by the environment; 2) ID of the current user goal; 3) ID of last user goal; 4) a scalar representation of student policy network's parameters under the current user goal; 5) a scalar representation of student policy network parameters under the last user goal.
\item The action $a_t$ corresponds user goal chosen $g_t$ by teacher policy model.
\item The reward $r$ consists of two parts, one is the reward $r_t^{or}$ from the \textit{Over-repetition Discriminator}, and the other $r_t^{c}$ is the change in episode total reward acquired by the student for the user goal $g_t$, formulated as:
\begin{eqnarray}
r_t =  r_t^{or} + r_t^{c} = r_t^{or} + x^{g_t}_t - x^{g_t}_{t'}
\end{eqnarray}
\end{itemize}

\begin{algorithm}
\caption{ ACL-DQN with Curriculum schedule A}  
\label{algA} 
\begin{algorithmic}[1]
	\State the DQN-based teacher model with probability $\epsilon$ select a random action $g_i$ in the user goal $G$;
	\State otherwise the DQN-based teacher model select $g_i = \mathop{\arg\max}_{g'}Q(s^t,g';\theta^T)$ in the user goal $G$;
\end{algorithmic} 
\end{algorithm}

\noindent where $x^{g_t}_{t'}$ is the previous episode total reward when the same user goal $g_t$ was trained on.

In this article, we user the deep Q-network (DQN) \cite{MnihKSRVBGRFOPB15} to improve the teacher policy based on teacher experience. In each step, the teacher agent takes the state $s_t$ as input and chooses the action $g_t$ to execute. The sampled user goal $g_t$ is handed over to the \textit{Over-repetition Discriminator} to judge whether it is over-sampling. if not, it will be passed to the user simulator as a goal to interact with the student agent, otherwise it will give the teacher agent a penalty. The more times the user goal has been selected, the greater penalty gave, the less the probability of being selected in the next step.
During training, we use $\epsilon$-greedy exploration that selects a random action with probability $\epsilon$ or otherwise follows the greedy policy $g_t = \mathop{\arg\max}_{g_t'}Q(s_t,g'_t;\theta^T)$. $Q(s_t,g_t; \theta^T)$ is the approximated value function, implemented as a Multi-Layer Perceptron (MLP) parameterized by $\theta^T$. 
When the dialogue terminates, the teacher agent then receives reward $r_t$, and updates the state to $s_{t+1}$. At each simulation epoch, we simulate $N$ ($N=1$) \footnote[1]{Considering the user cost in real dialogue scenarios, we set up only 1 simulation epoch for experience storage, $N=1$, to better reflect the performance of the proposed method on real dialogue tasks.} dialogues and store the experience $(s_t, g_t, r_t, s_{t+1})$ in the teacher experience replay buffer $D^T$ for teacher reinforcement learning. This cycle continues until the num\_episodes is reached.

\subsubsection{Curriculum schedule A}

As shown in Figure~\ref{curriculum A}, in order to evaluate the effect of a single DQN-based teacher model clearly, we replace the traditional sample method in user simulators with a single DQN-based teacher model that directly selects a user goal from the user goal set and dynamically adjust it according to the learning progress of the student agent and the over-repetition penalty using a $\epsilon$-greedy exploration (Algorithm~\ref{algA}).

\subsubsection{Curriculum schedule B}

In our curriculum schedule B, we make the learning process of the student agents similar to the education process of human students,  which is that students usually learn many easier curriculums before they start to learn more complex curriculums \cite{RenDLC18}.
Accordingly, we integrate user goal ranking in Curriculum schedule A, which allows student agents under the guidance of Curriculum schedule B to achieve progressive learning from easiness to hardness in proportion (Figure~\ref{curriculum B}).

We take the total number of inform\_slot and request\_slot $n$ in the user goals as a measure of the difficulty of each user goal.
According to this measure, user goals are divided into three groups from easiness to hardness: simple user goal set $G_{simple}$, medium user 

\begin{algorithm}
\caption{ ACL-DQN with Curriculum schedule B}  
\label{algB} 
\begin{algorithmic}[1]
\State Get the total number of inform\_slot $n_i$ and the number of request\_slot $n_r$ of each user goal, $n = n_i + n_r$;
\State Sort user goal set $G$ based on $n$ and divide it into three groups, simple user goal set $G_{simple}$ (30), medium user goal set $G_{medium}$ (72), and difficult user goal set $G_{difficult}$ (26);
\State Initialize curriculum\_phase = 'simple'
\If { $len(G_{curriculum\_phase)}/ len(G) * epoch\_size$ have been reached}
	\State curriculum\_phase = next\_difficult\_stage(); 
\Else
	\State curriculum\_phase = stay\_current\_stage();
\State the DQN-based teacher model with probability $\epsilon$ select a random action $g_i$ in $G_{curriculum\_phase}$;
\State otherwise the DQN-based teacher model select $g_i = \mathop{\arg\max}_{g'}Q(s^t,g';\theta^T)$ in $G_{curriculum\_phase}$; 
\EndIf
\end{algorithmic} 
\end{algorithm}

\noindent goal set $G_{medium}$, and difficult user goal set $G_{difficult}$. In the learning process of the student agents, we set the three user goal sets (from easiness to hardness) as the action set of the teacher agent sequentially to guarantee that the student agents learn the user goals of each stage in an orderly manner (Algorithm~\ref{algB}).

\subsubsection{Curriculum schedule C}

The curriculum schedule B may slow down the student agent learning. The reason is that even if the student agent has quickly mastered the goals of the current difficulty, it still needs to continue learning the remainder of this current difficulty. Accordingly, we design the curriculum schedule C, which is integrated "mastery" in curriculum schedule B, as shown in Figure~\ref{curriculum C}.  The curriculum schedule C supports the student agent to directly enter the user goal of the next stage without learning the remainder of the current difficulty if it has mastered the goals of the current difficulty.

It is considered that the student agent has mastered the user goals of this difficulty, if and only if the success rate of sampled user goals in the current difficulty exceeds the mastering threshold $\alpha (\alpha=0.5)$\footnote[2]{ We verified it in the subsequent experiment, the ACL-DQN performs best when the mastery threshold is 0.5.} within a continuous-time $T$ (T=5). 
The success rate of the sampled user goal in the current difficulty is $p_{success} = n_{success} / N_{sampled}$, where $n_{success}$ is the number of user goals completed by the student agent in the current difficulty, $N_{sampled}$ is the number of user goals sampled at the current difficulty (Algorithm~\ref{algC}).

\subsection{Over-repetition Penalty}

Under the three curriculum schedules mentioned above, the teacher policy model may  \textit{cheat} to obtain positive rewards, which is repeatedly selecting user goals that the student agent has mastered. Besides, it is clear that the limited size of replay memory makes overtraining even worse \cite{de2015importance}. Therefore, if the student agent is only restricted to some user goals already mastered, it will cause student

\begin{algorithm}
\caption{ ACL-DQN with Curriculum schedule C}  
\label{algC} 
\begin{algorithmic}[1]
\State Initialize curriculum\_phase = 'simple', a mastery threshold $\alpha$, a list $L$ for storing the success rate of the sampled user goal in the current difficulty;
	\State $p_{success} = n_{success} / N_{sampled}$;
	\State $L.append(p_{success})$
	\If {$episode \geq T$}
		\State $L.remove(0)$
	\EndIf
	\For {i in len(L)}
	\If { $L[i]$ $\geq$ $\alpha$ }
		\State $n = n + 1$
	\EndIf
	\EndFor
	\If {$n$ $\geq$ T}
		\State curriculum\_phase = next\_difficult\_stage();
	\Else 
		\State curriculum\_phase = stay\_current\_stage();
	\EndIf
\State the DQN-based teacher model with probability $\epsilon$ select a random action $g_i$ in $G_{curriculum\_phase}$;
\State otherwise the DQN-based teacher model select $g_i = \mathop{\arg\max}_{g'}Q(s^t,g';\theta^T)$ in $G_{curriculum\_phase}$;
\end{algorithmic} 
\end{algorithm}

\noindent agent learning to stagnate. For the sake of generalization of the proposed ACL-DQN method, we take into account guarantee the diversity of sampled user goals and integrate the over-repetition penalty mechanism in the framework.

Similar to the coverage mechanism in neural machine translation \cite{TuLLLL16}, we introduced an over-repetition vector $[og_1, og_2,...,og_n]$ to the teacher experience replay buffer $D^T$ for recording the sample times of each user goal. In the beginning, we initialize it as a zero vector with dimension $[1*n]$, where n is the number of user goals in the current user goal set. In each simulation training step, if a user goal $g_i$ is sampled, the corresponding variable over-repetition number $og_i$ is update by $og_i = og_i + 1$. The more times the user goal has been selected, the greater the over-repetition penalty gave by the over-repetition discriminator, the less the probability of being selected in the next step. Thus, an over-repetition penalty function $ORP(og)$ satisfies the following requirements:

\begin{itemize}
\item $ORP(og) \rightarrow [-L, 0]$.

\item $ORP(og)$ is a monotonically decreasing function of $og$.
\end{itemize}

\noindent where $L (L=40)$ is the maximum length of a simulated dialogue.

\subsection{Automatic Curriculum Learning}

The goal of student agents is to achieve a specific user goal through a sequence of actions with a user simulator, which can be considered as an MDP. In this stage, we use the DQN method to learn the student dialogue policy based on experiences stored in the student experience replay buffer $D^S$ :
\begin{itemize}
\item The state $s_t$ consists of five components: 1) one-hot representations of the current user action and mentioned slots; 2) one-hot representations of last system action and mentioned slots; 3) the belief distribution of possible value for each slot; 4) both a scalar and one-hot representation of current turn number; and 5) a scalar representation indicating the number of results which can be found in the database according to current search constraints.
\item The action $a_t$ corresponds pre-defined action set, such as request, inform, confirm\_question, confirm\_answer, etc.
\item The reward $r$: once a dialogue reaches the successful, the student agent receives a big bonus $2L$. Otherwise, it receives $-L$. In each turn, the student agent receives a fixed reward -1 to encourage shorter dialogues.
\end{itemize}

At each step, the student observes the dialogue $s$, and choose an action $a$, using an $\epsilon$-greedy.
The student agent then receives reward $r$,and updates the state to $s'$. Finally, we store the experience tuple $(s, a,r,s')$ in the student experience replay buffer $D^S$. This cycle continues until the dialogue terminates.

We improve the value function $Q(s, a,\theta^S)$ by adjusting $\theta^S$ to minimize the mean-squared loss function as follows:

\begin{eqnarray}
\begin{aligned}
&\mathcal{L}(\theta^S) = \mathbb{E}_{(s,a,r,s')\sim D^S}[(y_i- Q(s, a;\theta^S))^2]
\\
&y_i= r + \gamma \max _{a'}Q'(s',a';\theta^{S'})
\end{aligned}
\end{eqnarray}

\noindent where $\gamma \in [0,1]$ is a discount factor, and $Q'(\cdot)$ is the target value function that is only updated periodically. $Q(\cdot)$ can be optimized through $\nabla_{\theta^S} \mathcal{L}(\theta^S)$ by back-propagation and mini-batch gradient descent.

\subsection{Teacher Reinforcement Learning}

The teacher's function $Q(\cdot)$ can be improved using experiences stored in the teacher experience replay buffer $D^T$. In the implementation, we optimize the parameter $Q^T$ w.r.t. the mean-squared loss:

\begin{eqnarray}
\begin{aligned}
&\mathcal{L}(\theta^T) = \mathbb{E}_{(s,g,r,s')\sim D^T}[(y_i- Q(s, g;\theta^T))^2]
\\
&y_i= r_t^{or} + r_t^{change} + \gamma \max _{g'}Q'(s',g';\theta^{T'}) 
\end{aligned}
\end{eqnarray}

\noindent where $Q'(\cdot)$ is a copy of the previous version of $Q(\cdot)$ and is only updated periodically and $\gamma \in [0,1]$ is a discount factor. In each iteration, we improve $Q(\cdot)$ through $\nabla_{\theta^T} \mathcal{L}(\theta^T)$ by back-propagation and mini-batch gradient descent.

\section{Experiments}

Experiments have been conducted to evaluate the key hypothesis of ACL-DQN being able to improve the efficiency and stability of DQN-based dialogue policies, in two settings: simulation and human evaluation.

\subsection{Dataset}
Our ACL-DQN was evaluated on movie-booking tasks in both simulation and human-in-the-loop settings. Raw conversational data in the movie-ticket booking task was collected via Amazon Mechanical Turk with annotations provided by domain experts. The annotated data consists of 11 dialogue acts and 29 slots. In total, the dataset contains 280 annotated dialogues, the average length of which is approximately 11 turns.

\subsection{Baselines}

To verify the efficiency and stability of ACL-DQN, we developed different version of task-oriented dialogue agents as baselines to compare with.

\begin{itemize}
\item The \textbf{DQN} agent takes the user goal randomly sampled by the user simulator for leaning \cite{GaoWPLL18}.
\item The proposed \textbf{ACL-DQN}($A$) agent takes the curriculum goal specified by the teacher model equipped with \textit{curriculum schedule A} for automatic curriculum learning (Alforithm~\ref{algA}).
\item The proposed \textbf{ACL-DQN}($B$) agent takes the curriculum goal specified by the teacher model equipped with \textit{curriculum schedule B} for automatic curriculum learning (Alforithm~\ref{algB}).
\item The proposed \textbf{ACL-DQN}($C$) agent takes the curriculum goal specified by the teacher model equipped with \textit{curriculum schedule C} for automatic curriculum learning (Alforithm~\ref{algC}).
\end{itemize}

\begin{table*}[thb]
\centering
\resizebox{1.88\columnwidth}{!}{
\centering
\normalsize\begin{tabular}{lccccccccccccc}
\toprule
\multirow{2}*{\textbf{Agent}}& \multicolumn{3}{c}{Epoch = 100}& \multicolumn{3}{c}{Epoch = 200}& \multicolumn{3}{c}{Epoch = 300}& \multicolumn{3}{c}{Epoch = 400}\\
\cline{2-13}
&Success&Reward&Turns&Success&Reward&Turns&Success&Reward&Turns&Success&Reward&Turns\\
\hline
DQN&0.4012&-6.48&31.24&0.5242&10.36&27.08&0.6448&26.17&24.40&0.6598&28.73&22.88\\
ACL-DQN(A)&0.4309&-2.92&31.25&0.6159&22.99&23.84&0.7064&35.23&21.06&0.7419&40.19&19.66\\
ACL-DQN(B)&0.4202&-3.97&30.78&0.5678&16.29&25.69&0.6673&30.12&21.92&0.7073&35.81&20.11 \\
ACL-DQN(C)&\textbf{0.5717}&15.92&27.36&\textbf{0.7253}&37.39&21.30&\textbf{0.7573}&45.28&18.57 &\textbf{0.8055}&49.05&17.22 \\
\bottomrule
\end{tabular}
}
\caption{Result of different agents at $epoch = \{ 100, 200, 300, 400 \}$. Each number is averaged over 5 turns, each run tested on 50 dialogues. Success: Evaluated at the same epoch (except one group: at epoch 100, ACL-DQN(B)), ACL-DQN(A), ACL-DQN(B), and ACL-DQN(C) all outperform DQN, where ACL-DQN(C) has the best performance and ACL-DQN(B) has the worst performance in three curriculum schedule. The best scores are labeled in bold.}\smallskip
\label{tab:table1}
\end{table*}

\subsection{Implementation}

For all the models, we use MLPs to parameterize the value networks $Q(\cdot)$ with one hidden layer of size 80 and $tanh$ activation.
$\epsilon$-greedy is always applied for exploration.
We set the discount factor $\gamma = 0.9$.
The buffer size of $D^T$ and $D^S$ is set to 2000 and 5000, respectively.
The batch size is 16, and the learning rate is 0.001.
We applied gradient clipping on all the model parameters with a maximum norm of 1 to prevent gradient explosion.
The target network is updated at the beginning of each training episode.
The maximum length of a simulated dialogue is 40 turns.
The dialogues are counted as failed, if exceeding the maximum length of turns.
For training the agents more efficiently, we utilized a variant of imitation learning, called Reply Buffer Spiking (RBS) \cite{LiptonGLLAD16} at the beginning stage to build a naive but occasionally successful rule-based agent based on the human conversational dataset.
We also pre-filled the real experience replay buffer $B^u$ with 100 dialogues before training for all the variants of agents.

\subsection{Simulation Evaluation}

\subsubsection{Main result}

\begin{figure}[t]
\centering
\includegraphics[width=0.9\columnwidth]{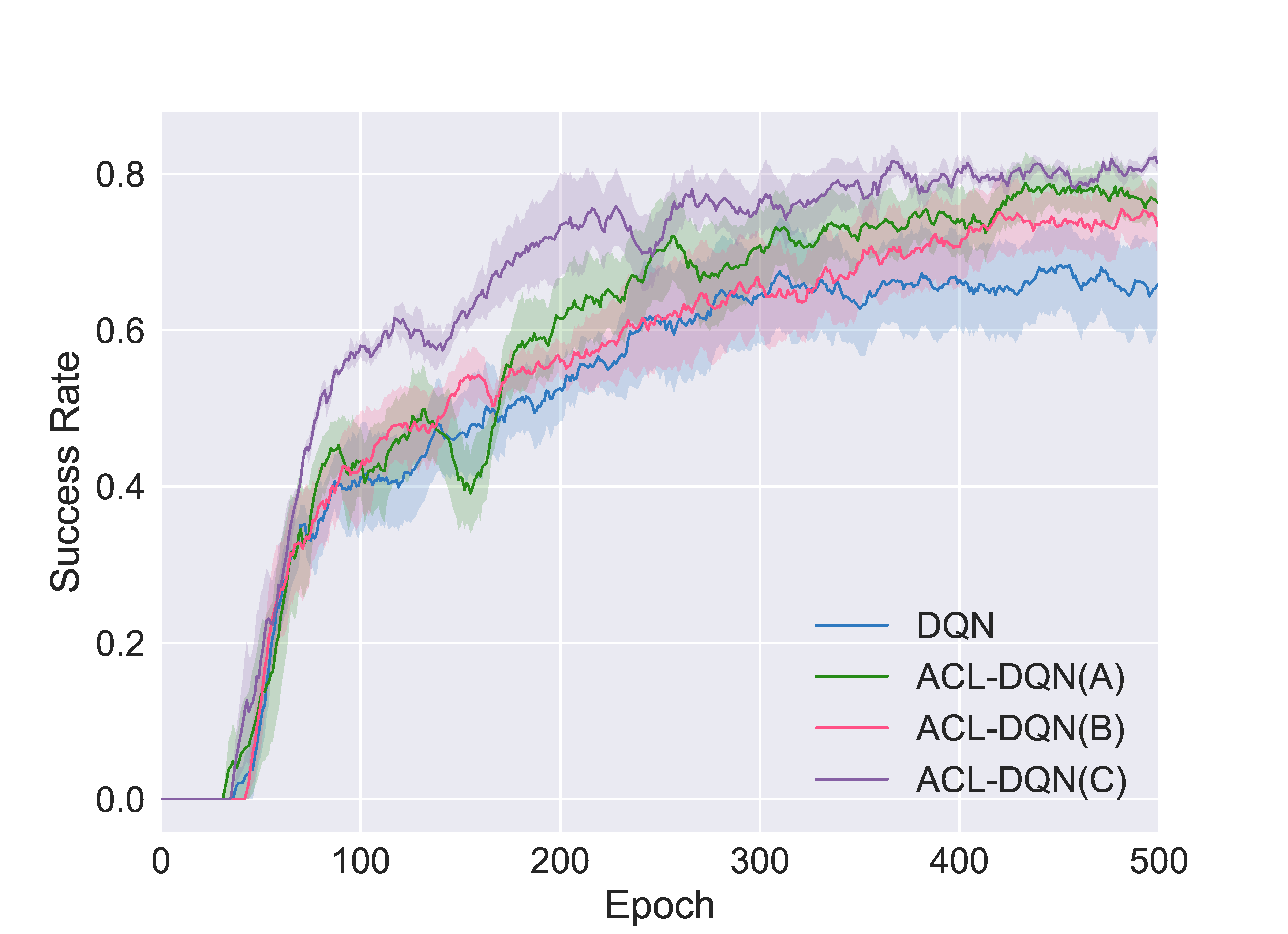} 
\caption{The learning curves of DQN, ACL-DQN(A), ACL-DQN(B), and ACL-DQN(C).}
\label{fig:main_result}
\end{figure}

\begin{figure}[tbp]
\centering
\includegraphics[width=0.9\columnwidth]{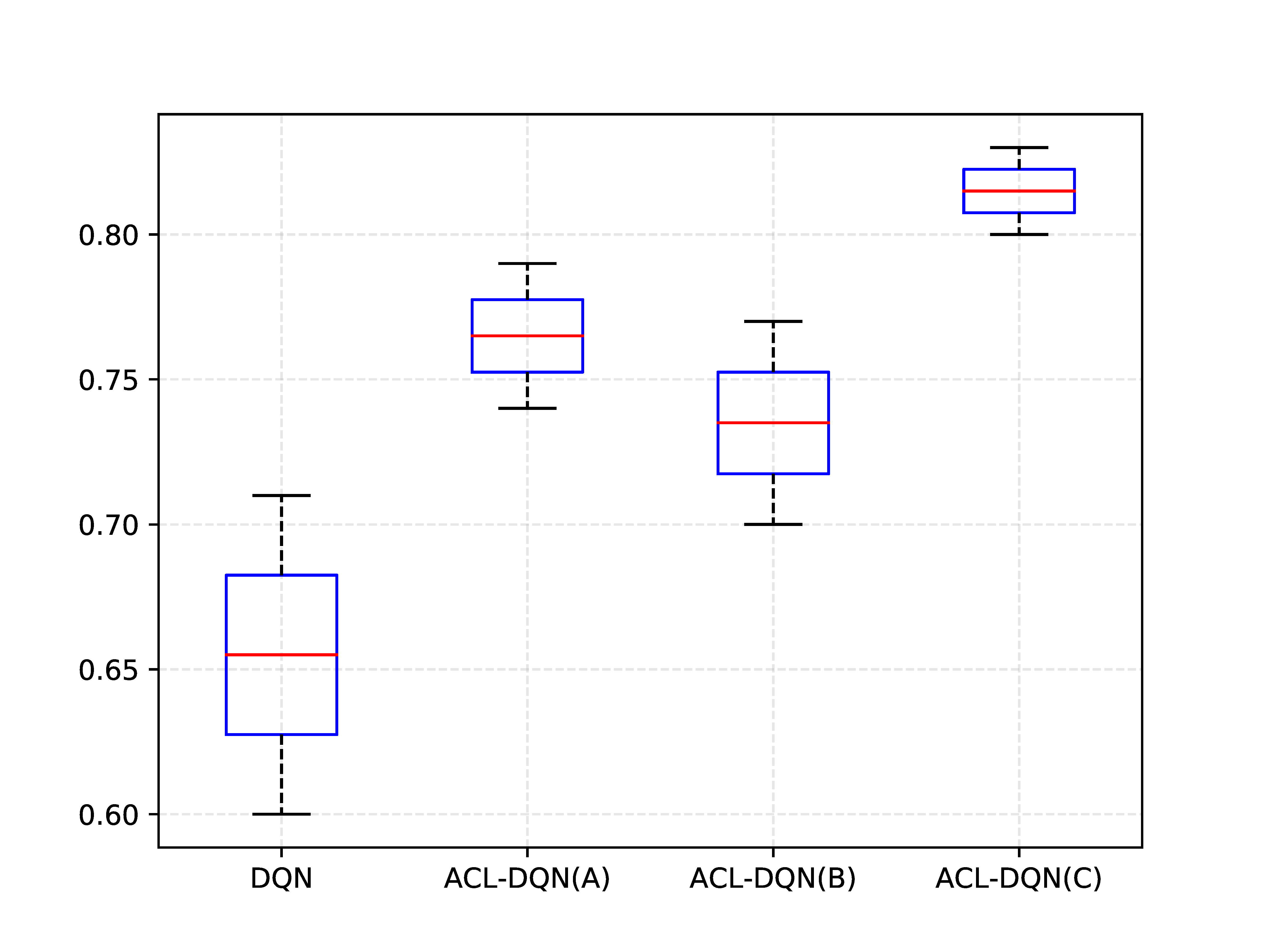} 
\caption{The stability of DQN, ACL-DQN(A), ACL-DQN(B), and ACL-DQN(C) about average success rate at 500 epoch.}
\label{fig:boxplot}
\end{figure}

The main simulation results are depicted in Table~\ref{tab:table1}, Figure~\ref{fig:main_result}, and~\ref{fig:boxplot}. The results show that all the ACL-DQN agents under three curriculum schedules significantly outperforms the baselines DQN with a statistically significant margin. Among them, ACL-DQN(C) shows the best performance, and ACL-DQN(B) shows the worst performance. The important reason is that, regardless of the mastering progress of the student agent and only let the student agent learning from easiness to hardness will slow down the learning of the student agent.
As shown in Figure~\ref{fig:main_result}, ACL-DQN(B) does not show significant advantages until after epoch $320$, while ACL-DQN(C) consistently outperform DQN by integrating the mastery module that monitors the learning progress of student agent and adjusts it in real-time.
Figure~\ref{fig:boxplot} is a boxplot of DQN and ACL-DQN under three curriculum schedules about the success rate at 500 epoch. It is clearly observed that ACL-DQN(A), ACL-DQN(B), and DQN-ACL(C) are more stable than DQN, where the average success rate of ACL-DQN(C) has stabilized above 0.8 while the DQN still fluctuates substantially around 0.65.
The result shows that ACL-DQN under the guidance of the teacher policy model shows a more effective and stable performance and the ACL-DQN(C) agent with the strength of supervision and controllability performs best and most stable.

\begin{figure*}
\centering
\subcaptionbox{DQN. \label{heat_mapa}}{\includegraphics[width=0.65\columnwidth]{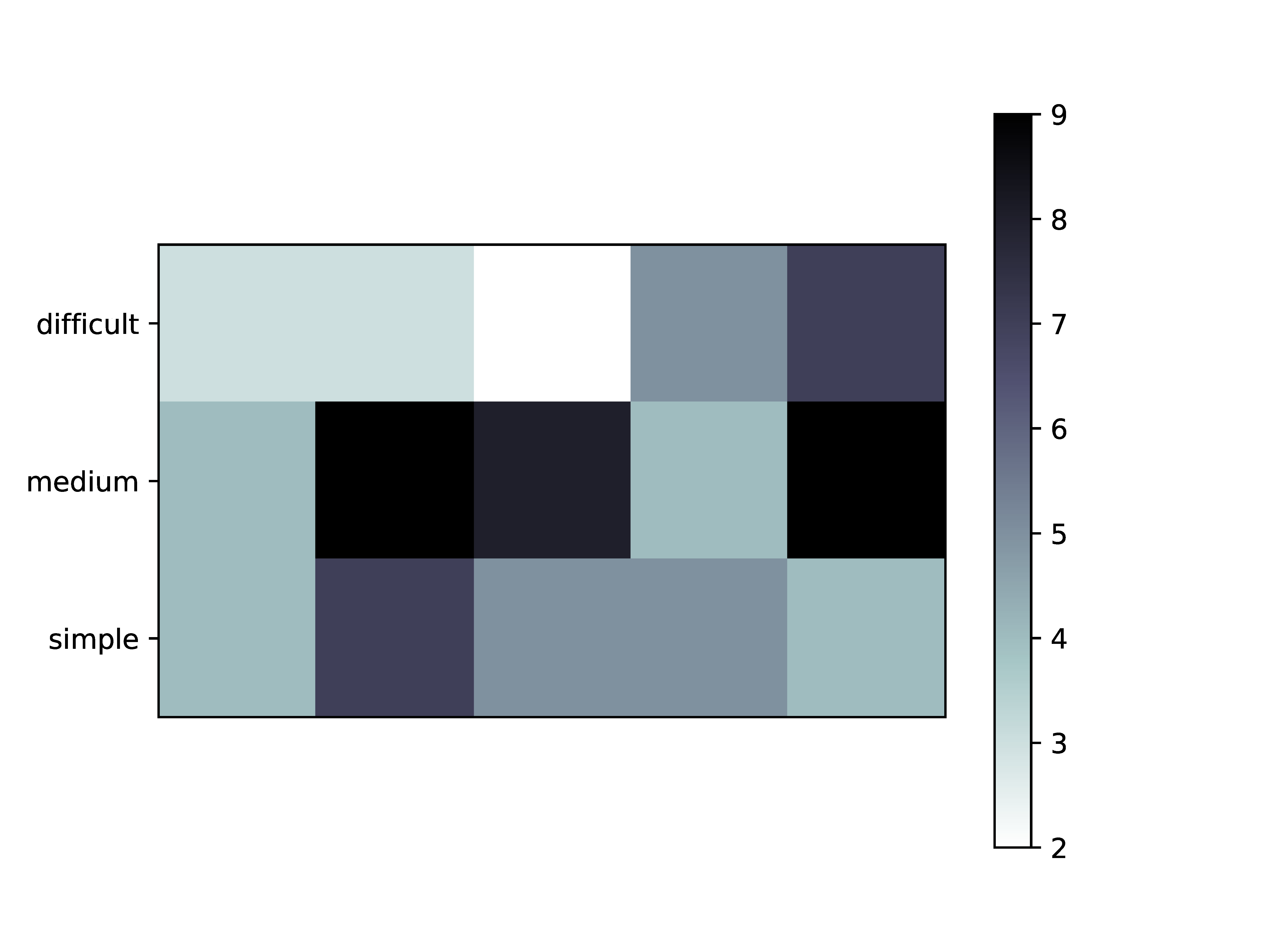}}
\subcaptionbox{ACL-DQN(A) w/o -ORP. \label{heat_mapb}}{\includegraphics[width=0.65\columnwidth]{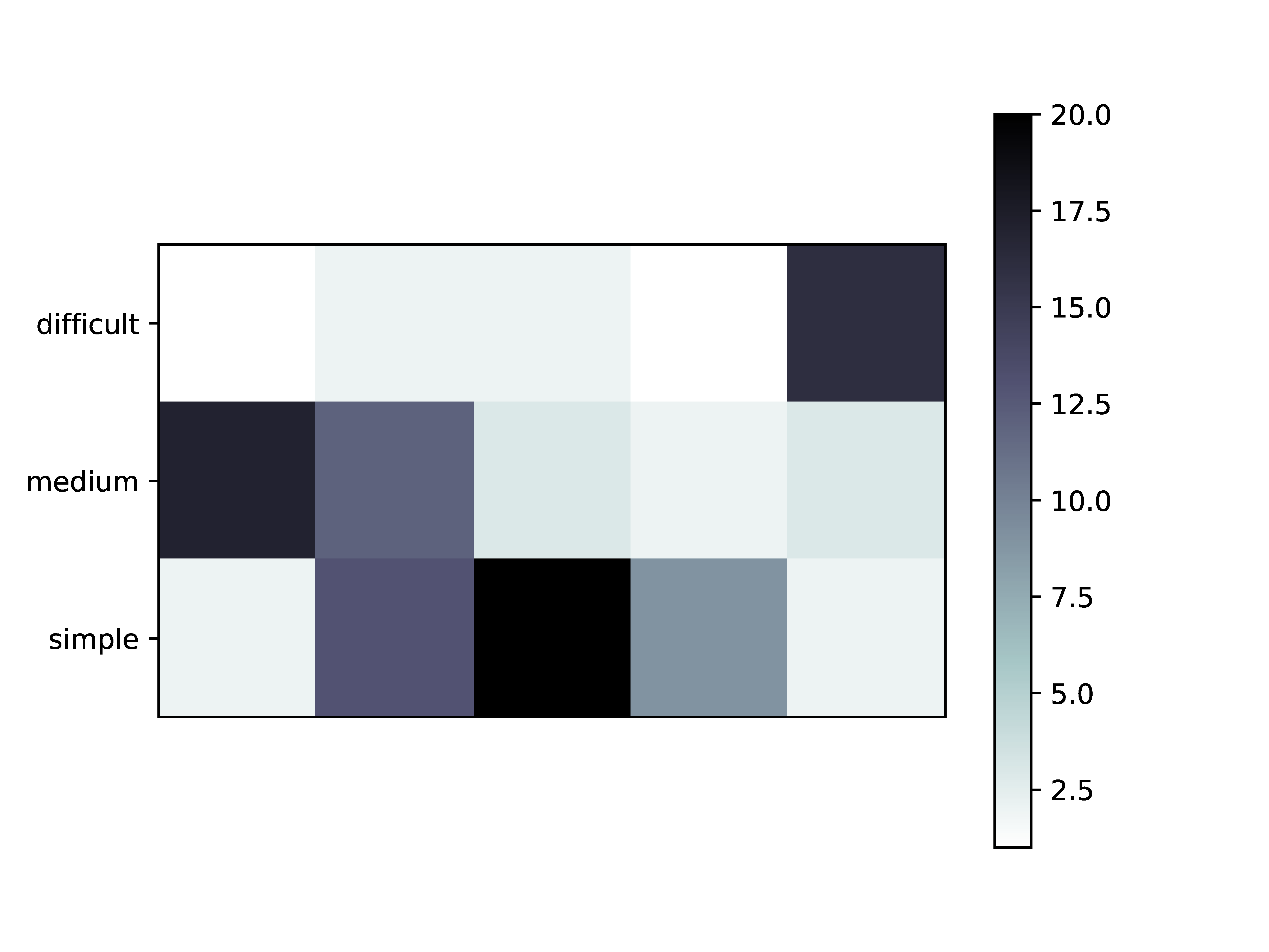}}
\subcaptionbox{ACL-DQN(A). \label{heat_mapc}}{\includegraphics[width=0.65\columnwidth]{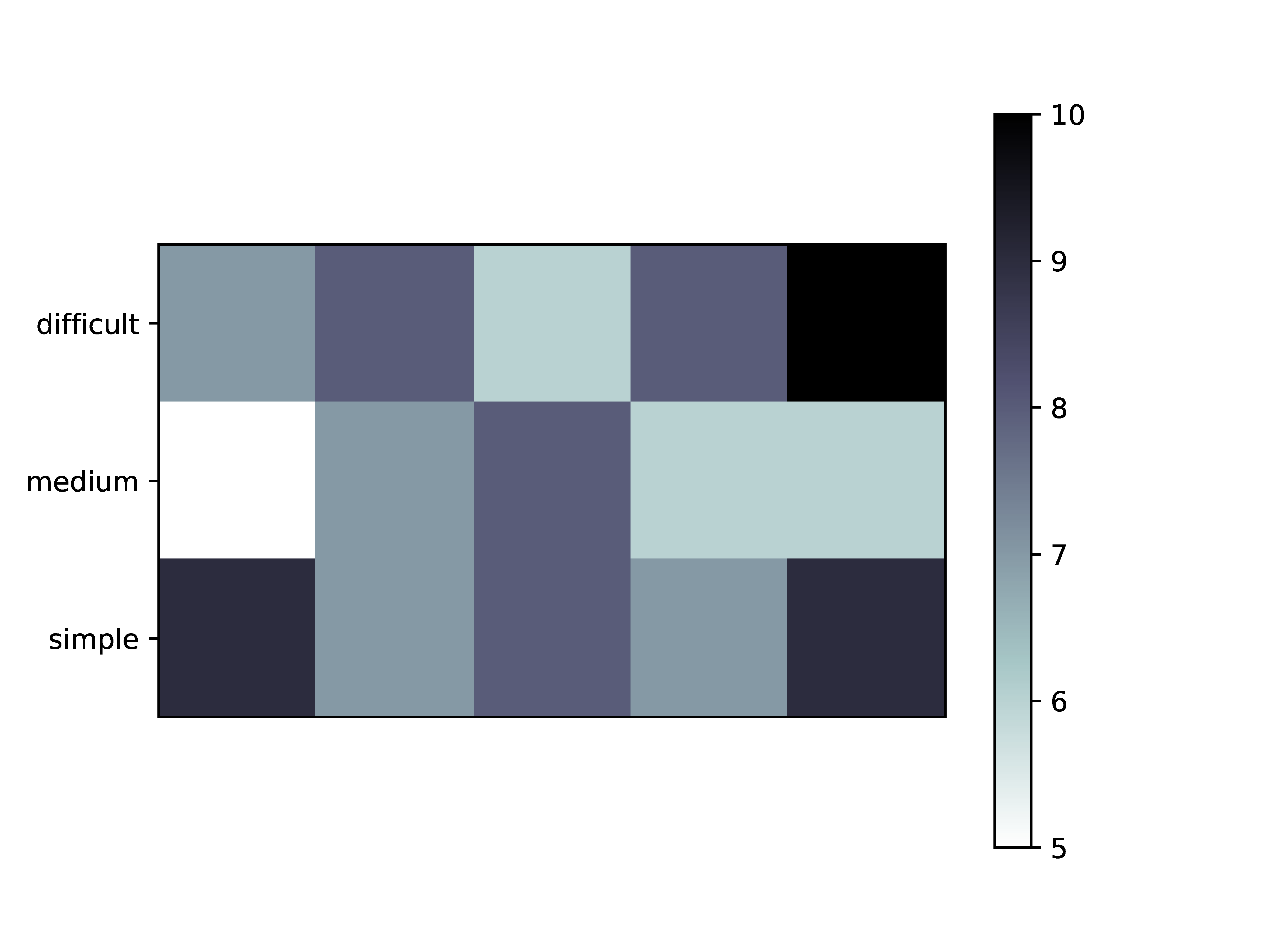}}
\caption{Heat maps of the number of the selected user goal in three different methods: (a) DQN, (b) ACL-DQN(A)/-ORP, (c) ACL-DQN(A). The depth of color in each image represents the number of times the goals has been select.}
\label{fig:heat_map}
\end{figure*}

\begin{figure}[t]
\centering
\includegraphics[width=0.9\columnwidth]{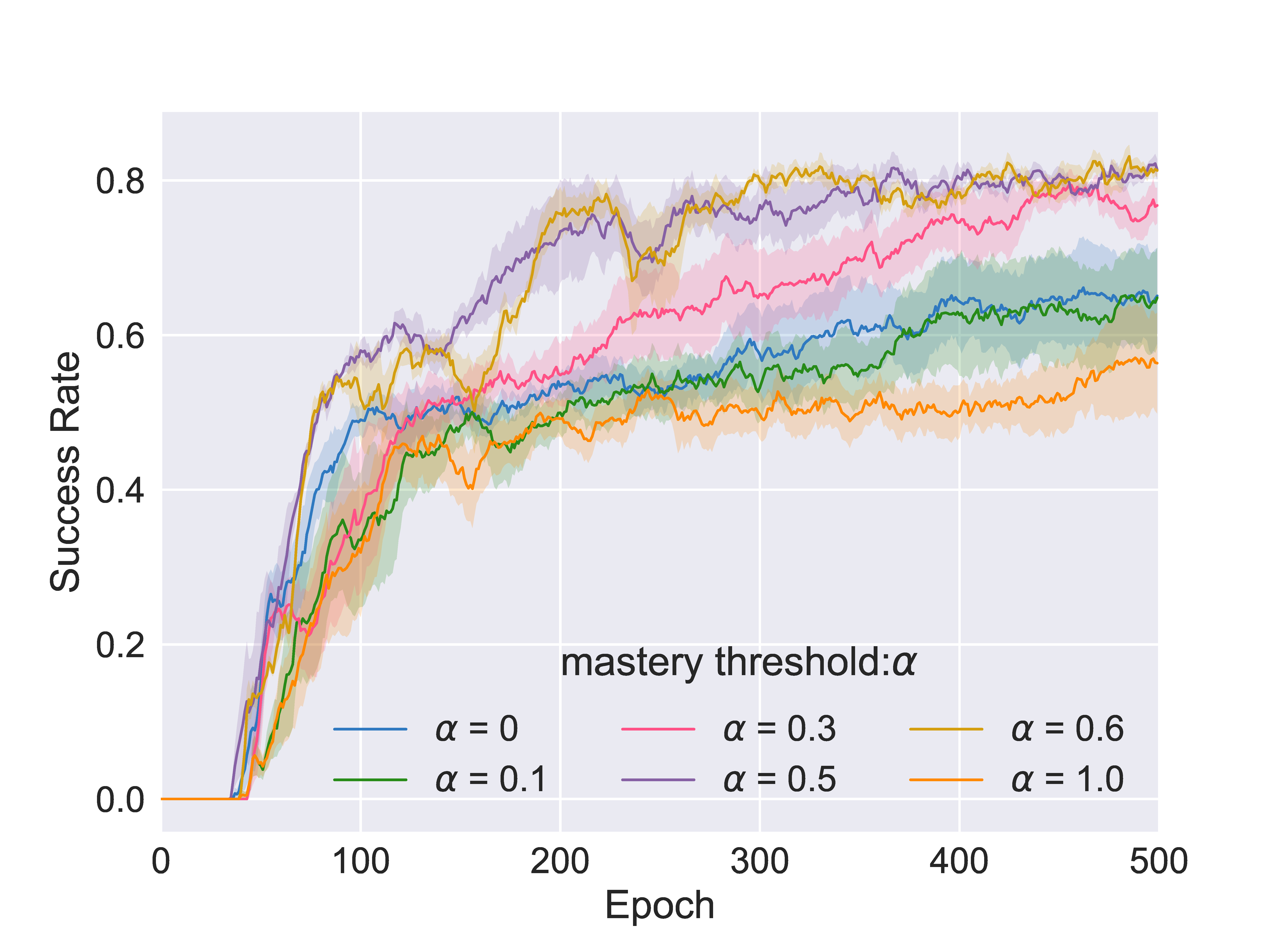} 
\caption{The \textit{mastery} in ACL-DQN(C): mastery threshold $\alpha$ in $[0.5,0.6]$ performs the best.}
\label{fig:mastery}
\end{figure}

\subsubsection{Mastery threshold of ACL-DQN(C)}

Choosing a new difficulty user goal set is allowed in ACL-DQN(C), if and only if the success rate of sampled user goals in the same difficulty has exceeded the "mastery" threshold within a continuous-time $T$ (details in Algorithm~\ref{algC}). Intuitively, if the threshold is too small, student agents will enter the learning of the harder goals before they mastered the simpler goals. The student agent is easy to collapse because it is difficult to learn positive training dialogues in time.
If the threshold is too big, the student agent will continue to learn the remaining simple goals even if they have mastered the simple goals, slowing down the efficiency of student agent learning.

Figure~\ref{fig:mastery} depicts the influences of different thresholds. As expected, when the threshold is too high or too small, it is difficult for student agents to lean a good strategy, and the learning rate of them is not as good as using a threshold within the range of $[0.5,0.6]$. The result here can serve as a reference to ACL-DQN(C) practitioners.

\subsubsection{Ablation Test}

To further examine the efficiency of the over-repetition penalty module, we conduct an ablation test by removing this module, referred to as ACL-DQN/-ORP. In order to observe the influence of the over-repetition penalty module more clearly, we take ACL-DQN(A) as an example to compare with ACL-DQN/-ORP and DQN with traditional randomly sampled.
We choose five user goals from the $C_{simple}$, $C_{medium}$ and $C_{difficult}$ and divide them into three groups according to their difficulty. The heat maps of three different methods (DQN, ACL-DQN(A)/-ORP, and ACL-DQN(A) ) are displayed in Figure~\ref{fig:heat_map}, where the color of grid reflects the number of the selected user goals. The darker the color, the more times the user goals have been selected. It is clear that the simple number in Figure~\ref{heat_mapa} is almost the same. But a serious imbalance phenomenon appears in Figure~\ref{heat_mapb}, which does ha

\subsection{Human Evaluation}

We recruited real users to evaluate different systems by interacting with different systems, without knowing which the agent is hidden from the users. At the beginning of each dialogue session,
the user randomly picked one of the agents to converse using a random user goal. The user can terminate the dialogue at any time if the user deems that the dialogue is too procrastinated and it is almost impossible to achieve their goal. Such dialogue sessions are considered as failed.
For the stability of different systems, each time the system was given a score (1-10), where the process was repeated 20 times. The greater the variance, the more unstable the system was. 

\begin{figure}[tbp]
\centering
\includegraphics[width=0.9\columnwidth]{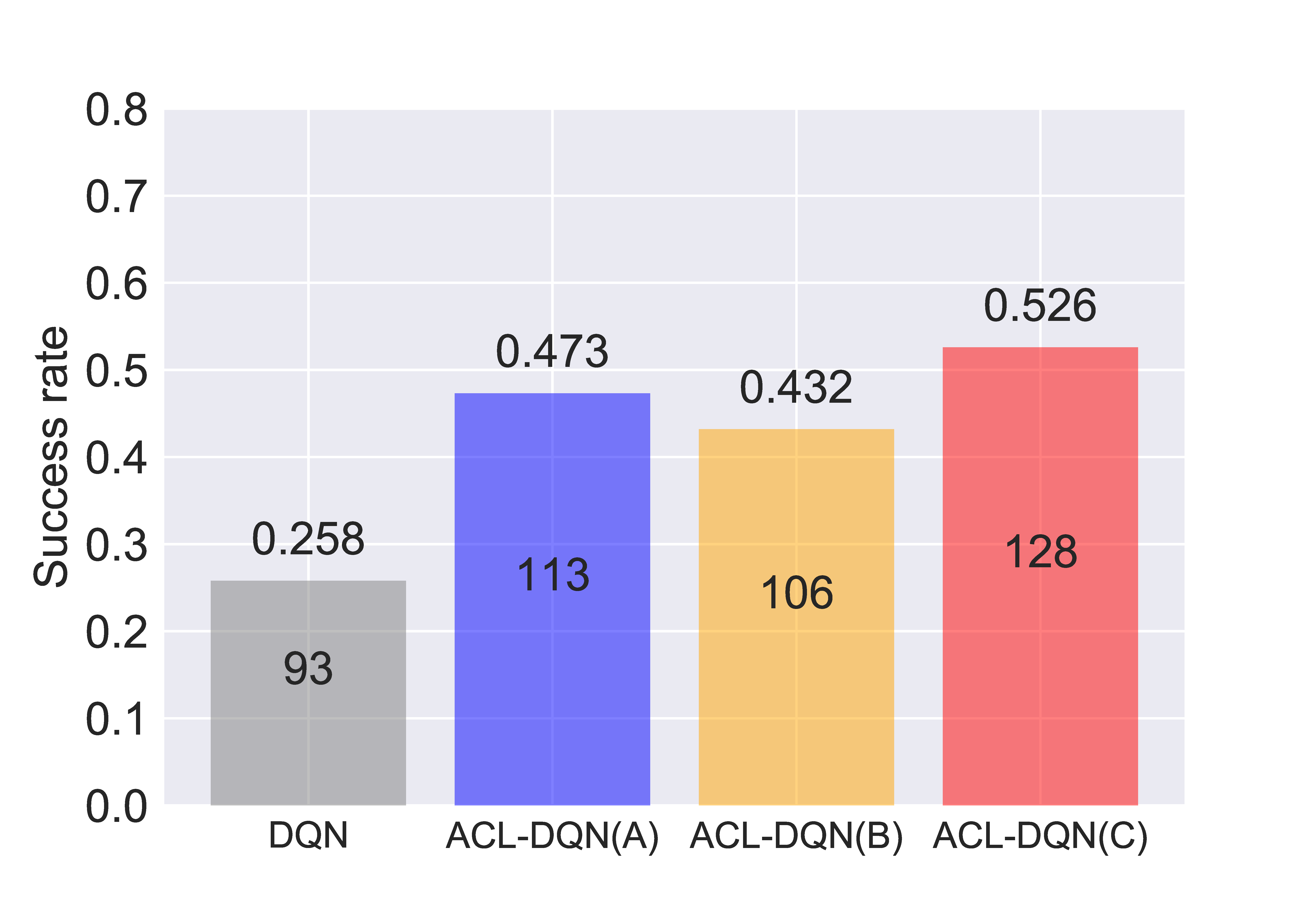} 
\caption{Human evaluation results of DQN, ACL-DQN(A), ACL-DQN(B), and ACL-DQN(C), the number of test dialogues indicated on each bar.}
\label{fig:human}
\end{figure}

Four agents (DQN, ACL-DQN(A), ACL-DQN(B), and ACL-DQN(C)) trained as previously described (Figure~\ref{fig:main_result}) at epoch 200 \textsuperscript{\rm 3}\footnotetext[3]{ Epoch 200 is picked since we are testing the efficiency of methods using a small number of real experiences.} are selected for human evaluation. As illustrated in Figure~\ref{fig:human}, the results of human evaluation confirm what we observed in the simulation evaluations. We find that DQN is abandoned more often due to its unstable performance, and it takes so many turns to reach a promising result in the face of more complex tasks (Figure~\ref{fig:main_result}), ACL-DQN(B) is kept not good enough since they could not adapt the harder goal quickly and the ACL-DQN(C) outperforms all the other agents. For the stability of different systems, the experimental results show that the variance of three ACL-DQN methods are all small than baselines, which means our methods are more stable, and ACL-DQN combined with the curriculum schedule C is the most stable one.

\section{Conclusion}
In this paper, we propose a novel framework, Automatic Curriculum Learning-based Deep Q-Network (ACL-DQN), to innovatively integrate curriculum learning and deep reinforcement learning in dialogue policy learning. We design a teacher model that automatically arranges and adjusts the sampling order of user goals without any requirement of prior knowledge to replace the traditional random sampling method in user simulators.
Sampling the user goals that match the ability of student agents regarding the difficulty of each user goal, maximizes and stabilizes student agents learning progress.
The learning progress of the student agent and the over-repetition penalty as the criteria of the sampling order of each user goal, guarantee both of the sampled efficiency and diversity.
The experimental results demonstrate the efficiency and stability of the proposed ACL-DQN.
Besides, the proposed method has strong generalizability, because it can be further improved by equipping with curriculum schedules.
In the future, we plan to explore the factors in the curriculum schedules that have a pivotal impact on dialogue policy learning, and evaluate the efficiency and stability of our approach by adopting different types of curriculum schedules.

\section{Acknowledgments.}
We thank the anonymous reviewers for their insightful feedback on the work, and we would like to acknowledge to volunteers from South China University of Technology for helping us with the human experiments. This work was supported by the Key-Area Research and Development Program of Guangdong Province,China (Grant No.2019B0101540042) and the Natural Science Foundation of Guangdong Province,China (Grant No.2019A1515011792).

\bibliography{Formatting-Instructions-LaTeX-2021}

\end{document}